# The Application of Bayesian Optimization and Classifier Systems in Nurse Scheduling

Jingpeng Li and Uwe Aickelin

School of Computer Science and Information Technology
The University of Nottingham
Nottingham, NG8 1BB, United Kingdom
{jpl, uxa}@cs.nott.ac.uk

**Abstract.** Two ideas taken from Bayesian optimization and classifier systems are presented for personnel scheduling based on choosing a suitable scheduling rule from a set for each person's assignment. Unlike our previous work of using genetic algorithms whose learning is implicit, the learning in both approaches is explicit, i.e. we are able to identify building blocks directly. To achieve this target, the Bayesian optimization algorithm builds a Bayesian network of the joint probability distribution of the rules used to construct solutions, while the adapted classifier system assigns each rule a strength value that is constantly updated according to its usefulness in the current situation. Computational results from 52 real data instances of nurse scheduling demonstrate the success of both approaches. It is also suggested that the learning mechanism in the proposed approaches might be suitable for other scheduling problems.

## 1 Introduction

Scheduling problems are generally NP-hard combinatorial problems, and a lot of research has been done to solve these heuristically ([2], [3], [8], [10]). However, research into the development of a general scheduling algorithm is still in its infancy.

Genetic Algorithms (GAs) ([6], [7]) mimicking the natural evolutionary process of the survival of the fittest, have attracted much attention in solving difficult scheduling problems in recent years. Some obstacles exist when using GAs: there is no canonical mechanism to deal with constraints, which are commonly met in most real-world scheduling problems, and small improvements of a solution are difficult. To overcome both difficulties, indirect approaches have been presented ([3], [9], [10]) for nurse scheduling and driver scheduling. In these indirect GAs, the solution space is mapped and then a separate decoding routine builds solutions to the original problem.

In our previous indirect GAs, learning was implicit and restricted to the efficient adjustment of weights for a set of rules that are used to construct schedules. The major limitation of those approaches is that they learn in a non-human way. Like most existing construction algorithms, once the best weight combination is found, the rules used in the construction process are fixed at each iteration. However, normally a long sequence of moves is needed to construct a schedule. Using fixed rules at each move is unreasonable and not coherent with human learning processes.

When a human scheduler works, he normally builds a schedule systematically following a set of rules. After much practice, the scheduler gradually masters the knowledge of which solution parts go well with others. He can identify good parts and is aware of the solution quality even if the scheduling process is not completed yet, thus having the ability to finish a schedule by using flexible, rather than fixed, rules. In this paper, we will present two more human-like scheduling approaches, by using a cutting-edge Bayesian Optimization Algorithm (BOA) and an Adapted Classifier System (ACS) individually, to implement explicit learning from past solutions.

In our test problem (nurse scheduling) problem, the number of the nurse is fixed (about 30), and the target is to create a weekly schedule by assigning each nurse one out of up to 411 shift patterns in the most efficient way. Both of the proposed approaches achieve this by using one suitable rule, from a rule set that contains a number of available rules, for each nurse's assignment. Thus, a potential solution is represented as a sequence of rules corresponding to the first nurse to the last nurse.

The long-term aim of our research is to model the learning of a human scheduler. Humans can provide high quality solutions, but this is tedious and time consuming. Typically, they construct schedules based on rules learnt during scheduling. Due to human limitations, these rules are typically simple. Hence, our rules will be relatively simple, too. Nevertheless, human generated schedules are of high quality due to the ability of the scheduler to switch between the rules, based on the state of the current solution. We envisage the proposed BOA and the ACS to perform this task.

## 2 The Nurse Scheduling Problem

Nurse scheduling has been widely studied recently ([4], [5]). The schedules generated have to satisfy working contracts and meet the demand for a given number of nurses of different grades on each shift. The problem is complicated by the fact that higher qualified nurses can substitute less qualified nurses but not vice versa. Thus scheduling the different grades independently is not possible. Due to this characteristic, finding and maintaining feasible solutions for most local search algorithms is difficult.

### 2.1 Integer Linear Programming

The nurse scheduling problem can be formulated as an Integer Program as follows:

Indices:
    $i = 1...n$ nurse index;
    $j = 1...m$ shift pattern index;
    $k = 1...14$ day and night index (1...7 are days and 8...14 are nights);
    $s = 1...p$ grade index.

Decision variables:

$x_{ij} = 1$ if nurse $i$ works shift pattern $j$ otherwise $x_{ij} = 0$.

Parameters:

$m$ = Number of shift patterns;
$n$ = Number of nurses;
$p$ = Number of grades;
$a_{jk} = 1$ if shift pattern $j$ covers day/night $k$ otherwise $a_{jk} = 0$;
$q_{is} = 1$ if nurse $i$ is of grade $s$ or higher otherwise $q_{is} = 0$;
$p_{ij}$ = Preference cost of nurse $i$ working shift pattern $j$;
$R_{ks}$ = Demand of nurses with grade $s$ on day/night $k$;
$N_i, D_i, B_i$ = Shifts per week of nurse $i$ if night / day / both shifts are worked;
$F(i)$ = Set of feasible shift patterns for nurse $i$, where $F(i)$ is defined as

$$F(i) = \begin{cases} \sum_{k=1}^{7} a_{jk} = D_i, & \forall j \in \text{day shifts} \\ \sum_{k=8}^{14} a_{jk} = N_i, & \forall j \in \text{night shifts} \\ \sum_{k=1}^{14} a_{jk} = B_i, & \forall j \in \text{combined shifts} \end{cases}, \forall i.$$

Target function is to minimize total preference cost of all nurses, denoted as

$$\sum_{i=1}^{n} \sum_{j \in F(i)}^{m} p_{ij} x_{ij} \to \min! \qquad (1)$$

Subject to:

1. Every nurse works exactly one feasible shift pattern:

$$\sum_{j \in F(i)} x_{ij} = 1, \forall i; \qquad (2)$$

2. The demand for nurses is fulfilled for every grade on every day and night:

$$\sum_{j \in F(i)} \sum_{i=1}^{n} q_{is} a_{jk} x_{ij} \geq R_{ks}, \forall k, s. \qquad (3)$$

Constraint set (2) ensures that every nurse works exactly one shift pattern from his/her feasible set, and constraint set (3) ensures that the demand for nurses is covered for every grade on every day and night. Note that the definition of $q_{is}$ is such that higher graded nurses can substitute those at lower grades if necessary.

Typical problem dimensions are $n = 30$ nurses of $p = 3$ grades and $m = 411$ shift patterns for each nurse. Thus, the integer programming has some 12000 binary variables and about 100 constraints. This is a moderately sized problem. However, some problem cases remain unsolved after overnight computation using professional software [4].

### 2.2 A Graphic Representation for Nurse Scheduling

Figure 1 shows a graphical representation of the solution structure of the problem: a hierarchical and acyclic directed graph. The node $N_{ij} (i \in \{1,2,...,n\}; j \in \{1,2,...,r\})$ in the graph denotes that nurse *i* is assigned by using rule *j*, where *n* is the number of nurses to be scheduled and *r* is the number of rules to be used in the building process. The directed edge (arrow) from node $N_{ij}$ to node $N_{i+1,j'}$ denotes a causal relationship of "$N_{ij}$ following $N_{i+1,j'}$", i.e. a rule sub-string for nurse *i* where the previous rule is *j* and the current rule is *j'*. In this graph, a possible solution (a complete rule string) is represented as a directed path from nurse 1 to nurse *n* connecting *n* nodes.

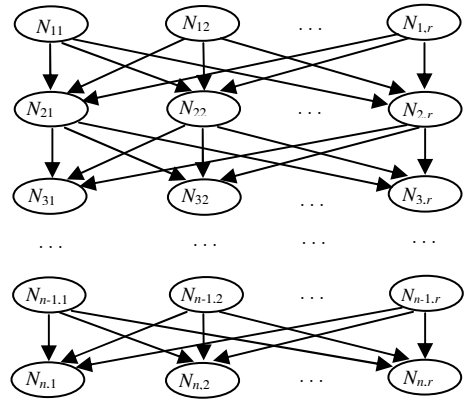

**Fig. 1.** A directed graph for nurse scheduling

## 3   A Building Heuristic for Nurse Scheduling

Similar to the human's working pattern, a building heuristic is designed to build a schedule step by step by using a set of rules. As far as the domain knowledge of nurse scheduling is concerned, the following four rules are currently applied.

The first rule, called '*Random*' rule, is used to select a nurse's shift pattern at random. Its purpose is to introduce randomness into the search thus enlarging the search space, and most importantly to ensure that the proposed algorithm has the ability to escape from local optimum. This rule mirrors much of a scheduler's creativeness to come up with different solutions if required.

The second rule is the '*k-Cheapest*' rule. Disregarding the feasibility of the schedule, it randomly selects a shift pattern from a *k*-length list containing patterns with *k*-cheapest cost $p_{ij}$, in an effort to reduce the cost of a schedule as much as possible.

The third rule '*Cover*' is designed to consider only the feasibility of the schedule. It schedules one nurse at a time to cover those days and nights with the highest number of uncovered shifts. For each shift in a nurse's feasible set, we calculate the total number of uncovered shifts that would be covered if the nurse worked that shift pattern. In order to ensure that high-grade nurses are not 'wasted' covering unnecessarily for lower-grade nurses, for nurses of grade *s*, only the shifts requiring grade *s* nurses are counted as long as there is a single uncovered shift for this grade. If all these are covered, shifts of the next lower grade are considered and once these are filled those of the next lower grade. Hence, the '*Cover*' rule can be summarised as finding those shift patterns with the largest amount of undercover of the highest feasible grade.

The fourth rule '*Contribution*' is biased towards solution quality but includes some aspects of feasibility by computing an overall score for each feasible pattern for the nurse currently being scheduled. It is designed to take into account the nurses' preferences. It also takes into account some covering constraints in which it gives preference to patterns that cover shifts that have not yet been allocated sufficient nurses to meet their total requirements. This is achieved by going through the entire set of feasible shift patterns for a nurse and assigning each one a score. The one with the highest (i.e. best) score is chosen. In formulation, the score of a shift pattern $S_{ij}$ is denoted as

$$S_{ij} = w_p(100 - P_{ij}) + \sum_{s=1}^{3} w_s q_{is} (\sum_{k=1}^{14} a_{jk} d_{ks}), \qquad (4)$$

where $w_p$ is the weight of the nurse's $p_{ij}$ value for the shift pattern, $w_s$ is the weight of covering an uncovered shift of grade $s$, and $d_{ks} = 1$ if there are still nurses needed on day $k$ of grade $s$ otherwise $d_{ks} = 0$.

Independent of the rules used, the fitness of completed solutions has to be calculated. Unfortunately, feasibility cannot be guaranteed. This is a problem-specific issue and cannot be changed. Therefore, we need a penalty function approach. Since the chosen encoding automatically satisfies constraint set (2), we can use the following formula to calculate the fitness of solutions:

$$\sum_{i=1}^{n} \sum_{j=1}^{m} p_{ij} x_{ij} + w_{demand} \sum_{k=1}^{14} \sum_{s=1}^{p} \max\left[ R_{ks} - \sum_{i=1}^{n} \sum_{j=1}^{m} q_{is} a_{jk} x_{ij} ; 0 \right] \to \min!, \qquad (5)$$

where $w_{demand}$ is the penalty weight. Note that the penalty is proportional to the number of uncovered shifts.

## 4 A Bayesian Optimization Algorithm

Bayesian networks [11] are often used to model multinomial data with both discrete and continuous variables by encoding the relationship between the variables contained in the modelled data. Thus, they represent the structure of a problem. Moreover, Bayesian networks can be used to generate new instances of the variables with similar properties as those given. Each node in the network corresponds to one variable, and each variable corresponds to one position in the strings representing the solutions. The relationship between two variables is represented by a directed edge between the two corresponding nodes (as seen in Figure 1).

Any complete probabilistic model of a domain must represent the joint distribution, i.e. the probability of every possible event as defined by the values of all the variables. The number of such events is exponential. To achieve compactness, Bayesian networks factor the joint distribution into local conditional distributions for each variable.

Mathematically, an acyclic Bayesian network encodes a full joint probability distribution by the product

$$P(x_1,...,x_n) = \prod_{i=1}^{n} P(x_i \mid pa(X_i)), \qquad (6)$$

where $x_i$ denotes some values of the variable $X_i$, $pa(X_i)$ denotes a set of values for parents of $X_i$ in the network (the set of nodes from which there exists an individual edge to $X_i$), and $P(x_i \mid pa(X_i))$ denotes the conditional probability of $X_i$ conditioned on variables $pa(X_i)$. This distribution can be used to generate new instances using the marginal and conditional probabilities.

### 4.1 Learning based on the Bayesian Network

The graph shown in Figure 1 can be regarded as a Bayesian network, which denotes the solution structure of the problem. In this network, learning the best rule sequence amounts to counting the frequency of using each rule. Hence, we use the symbol '#' meaning 'the number of' in the following equations. It calculates the conditional probabilities of each possible value for each node given all possible values of its parents. For example, for node $N_{i+1,j'}$ with a parent $N_{ij}$, its conditional probability is

$$P(N_{i+1,j'} \mid N_{ij}) = \frac{P(N_{i+1,j'}, N_{ij})}{P(N_{ij})} = \frac{\#(N_{i+1,j'} = true, N_{ij} = true)}{\#(N_{i+1,j'} = true, N_{ij} = true) + \#(N_{i+1,j'} = false, N_{ij} = true)}. \qquad (7)$$

Note that nodes $N_{1j}$ have no parents. In this circumstance, their probabilities are

$$P(N_{1j}) = \frac{\#(N_{1j} = true)}{\#(N_{1j} = true) + \#(N_{1j} = false)} = \frac{\#(N_{1j} = true)}{T}. \qquad (8)$$

These probability values can be used to generate new rule strings, or new solutions. Since the first rule in a solution has no parents, it will be chosen from nodes $N_{1j}$ according to their probabilities. The next rule will be chosen from nodes $N_{ij}$ according to their probabilities conditioned on the previous nodes. This building process is repeated until the last node has been chosen from nodes $N_{nj}$, where $n$ is number of the nurses. A link from nurse 1 to nurse $n$ is thus created, representing a new possible solution. Since all the probability values are normalized, the roulette-wheel method is a good strategy for rule selection.

### 4.2 A BOA Approach for Nurse Scheduling

The BOA is applied to learn good partial solutions and then to complete them by building a Bayesian network of the joint distribution of solutions [12]. The nodes, or variables, in the Bayesian network correspond to the individual rules from which a schedule will be built step by step. In the proposed BOA, the first population of rule strings is generated at random. From the current population, a set of better rule strings is selected. Any selection method biased towards better fitness can be used, and in this paper, the traditional roulette-wheel selection is applied. The conditional probabilities of each node in the Bayesian network are computed. New rule strings are generated by using these conditional probability values, and are added into the old population, replacing some of the old rule strings. In detail:
1. Set $t = 0$, and generate an initial population $P(0)$ at random;
2. Use roulette-wheel to select a set of promising rule strings $S(t)$ from $P(t)$;

3. Compute the conditional probabilities of each node according to this set of promising solutions;
4. For each nurse's assignment, use the roulette-wheel method to select one rule according to the conditional probabilities of all available nodes, thus obtaining a new rule string. A set of new rule strings $O(t)$ will be generated in this way;
5. Create a new population $P(t+1)$ by replacing some rule strings from $P(t)$ with $O(t)$, and set $t = t+1$;
6. If the termination conditions are not met, go to step 2.

## 5   An Adapted Classifier System

The classifier system is an induction self-learning system in which a set of condition-action rules, called classifiers, compete to control the system and gain credit based on the system's receipt of reinforcement from the environment. It was first introduced by Holland in 1975 and has been extensively studied by others in recent years. In original classifier systems [7], the learning procedures consist of two parts: *credit assignment* and *rule discovery*. The former is critical, which is achieved by using a "bucket brigade" algorithm to rate the rules the system already has. The latter is applied very seldom, which is achieved by using GAs to replace rules of low strength and provide new rules when environmental situations are ill handled.

The design of our ACS is based on the idea of learning from the environment by providing the system with some measure of its performance. In particular, we study reinforcement learning for entities (i.e. for each nurse/rule combination shown in Figure 1). Each entity is given a learning task and when all tasks are completed, a solution is built. This solution will then receive a positive or negative reward according to its quality. The reward is shared among all entities involved. Thus, the process is similar to a game such as chess, where many moves are made before feedback is received.

In this approach, each building unit has its strength showing its usefulness in the current situation, and this strength is constantly assessed and updated. To implement learning based on previous solutions, an ACS for nurse scheduling is designed, which consists of the following four steps:

1. Initialise the strengths of all nodes in Figure 1 by assigning each node a same constant value, and create an initial solution by randomly picking a rule from the rule set for each nurse's assignment;
2. Considering the strengths of all nodes in the graph, we use the roulette-wheel method to select one node for each nurse, i.e. selection is biased towards higher strength. New solutions are generated in this way;
3. If a new solution is better than the previous one, a positive reward is received by this solution and evenly assigned to every associated node, otherwise a negative reward is received and evenly assigned to associated nodes;
4. Keep the best solution found so far. If ending conditions (maximum number of iterations) are not met, go to step 2.

To help understanding how the reward is assigned and shared, we will give a simple example of scheduling three nurses using four rules. The initial strength of each node is set to 10, and the reward of an improved solution is set to 3. The initial solution is generated by using rule 1 for nurse 1, rule 4 for nurse 2 and rule 3 for nurse 3. The next solution is generated by using rule 4 for nurse 1, rule 2 for nurse 2 and rule 3 for nurse 3. Thus, the strength matrix after each generation is updated as follows:

$$\begin{pmatrix} 10 & 10 & 10 & 10 \\ 10 & 10 & 10 & 10 \\ 10 & 10 & 10 & 10 \end{pmatrix} \Rightarrow \begin{pmatrix} 11 & 10 & 10 & 10 \\ 10 & 10 & 10 & 11 \\ 10 & 10 & 11 & 10 \end{pmatrix} \Rightarrow \begin{pmatrix} 11 & 10 & 10 & 11 \\ 10 & 11 & 10 & 11 \\ 10 & 10 & 12 & 10 \end{pmatrix}$$

It is worth mentioning here, that our proposed two approaches in the way of building schedules may have similarity with ant colony optimisation. However, their search mechanisms are very different. In our BOA, the search is based on the conditional probabilities of all available moves, rather than on the local and global trail updating in the ants' method. In our ACS, the searching method is still under development. It is currently based on the improvement of a single path, rather than the evolution of a group of paths in ant algorithms.

## 6   Computational Results

In this section, we present the results of extensive computer experiments on 52 real data instances and compare them to results of the same data instances found previously by other algorithms. Figure 2 summarises results of 20 runs with different random seeds for the BOA and the ACS respectively. Figure 3 gives an overall comparison between various algorithms. The runtime of both algorithms is approx 10-20 seconds per run and data instance on a Pentium 4 PC.

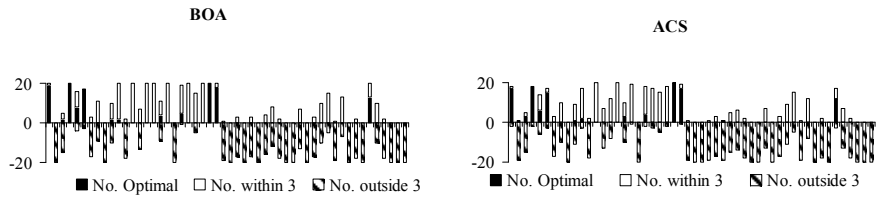

**Fig. 2.** Results of the BOA and the ACS

In Figure 2, the *x*-axis represents the number of data sets, and the bars above the *y*-axis represent solution quality. The black bars show the number of optimal, the white near-optimal (within three units) solutions. The bars below the y-axis represent the number of times the algorithm failed to find a feasible solution or the solution was feasible but non-optimal (more than three units from optimum). The value of three units was chosen in consultation with the hospital involved. Hence, the shorter the bar below the y-axis and the longer above, the better the algorithm's performance.

Figure 2 shows that for the BOA 38 out of 52 data sets are solved to or near to optimality. Additionally, feasible solutions are always found for all data sets. Broadly speaking, the results for the ACS are similar, but a little weaker. This is unsurprising as in its present form the ACS is simple and its search is based on a single solution.

Figure 3 gives the optimal or best-known solutions found by an IP software package, and compares performance of different GAs ([1], [3]) with the BOA and the ACS presented here. The results are encouraging, with a fraction of the development time and simpler algorithms, the complex genetic algorithms are outperformed in terms of feasibility, best and average results. Only the hill-climbing GA, which includes an additional local search, has a better 'best case' performance. We believe that once this feature is added into our approach, by using the ACS as the hill-climber for the BOA, we will see the best possible results. Our plan is to implement a post-processor that is similar to a human scheduler who 'improves' a finished schedule.

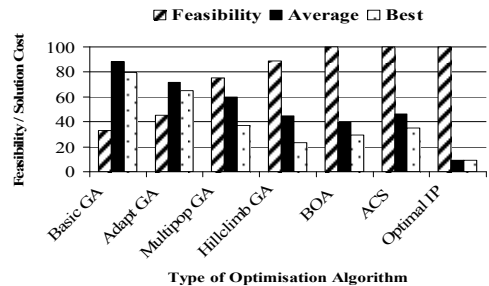

**Fig. 3.** Summary results of various algorithms

Another direction for further research is to see if there are good constructing sequence for a fixed nurses' scheduling order. If so, the good patterns could be recognized and then extracted as new domain knowledge. Then using the extracted knowledge, we can assign specific rules to the corresponding nurses beforehand, and only need to schedule the remaining nurses, hence reducing the solution space.

## 7 Conclusions

This paper presents two scheduling algorithms based on the Bayesian optimization and classifier systems. The approach is novel because it is the first time that ideas from classifier systems are applied to personnel scheduling. Two effective methods are proposed to implement explicit learning from past solutions. Unlike most existing approaches, the new approach has the ability to build schedules using flexible, rather than fixed rules. Experimental results from real-world nurse scheduling problems demonstrate the strength of the proposed approaches.

Although we have presented this work in terms of nurse scheduling, it is suggested that the main ideas of the approaches could be applied to many other scheduling problems where the schedules will be built systematically according to specific rules. It is also hoped that this research will give some preliminary answers about how to include human-like learning into scheduling algorithms and may therefore be of interest to practitioners and researchers in areas of scheduling and evolutionary computation. In future, we aim to extract the 'explicit' part of the learning process further, e.g. by keeping learnt rule sequences from one data instances to the next.

**Acknowledgements**

The work was funded by the UK's, Engineering and Physical Sciences Research Council (EPSRC), under grand GR/R92899/01.

**References**

1. Aickelin, U., Dowsland, K.: Exploiting Problem Structure in a Genetic Algorithm Approach to a Nurse Rostering Problem. Journal of Scheduling 3(2000) 139-153
2. Aickelin, U., Dowsland, K.: Enhanced Direct and Indirect Genetic Algorithm Approaches for a Mall Layout and Tenant Selection Problem. Journal of Heuristics 8(2002) 503-514
3. Aickelin, U., Dowsland, K.: An Indirect Genetic Algorithm for a Nurse Scheduling Problem. Computers and Operations Research 31(2003) 761-778
4. Aickelin, U., White P.: Building Better Nurse Scheduling Algorithms. Annals of Operations Research 128 (2004) 159-177
5. Burke, E.K., Cowling, P.I., Causmaecker, P., Vanden Berghe, G.: A Memetic Approach to the Nurse Rostering Problem. In: Applied Intelligence, Vol. 15. Kluwer (2001) 199-214
6. Goldberg, D.E.: Genetic Algorithms in Search, Optimization and Machine Leaning. Addison-Wesley (1989)
7. Holland, J.H.: Adaptation in Natural and Artificial Systems. University of Michigan Press (1975), Republished by MIT Press (1992)
8. Li, J., Kwan, R.S.K.: A Fuzzy Simulated Evolution Algorithm for the Driver Scheduling Problem. In: Proceedings of Congress on Evolutionary Computation. (2001a) 1115-1122
9. Li, J., Kwan, R.S.K.: A Fuzzy Theory Based Evolutionary Approach for Driver Scheduling. In: Spector, L. et al. (eds.): Proceedings of Genetic and Evolutionary Computation Conference (GECCO). Morgan Kaufmann Publishers (2001b) 1152-1158
10. Li, J., Kwan, R.S.K.: A Fuzzy Genetic Algorithm for Driver Scheduling. European Journal of Operational Research 147 (2003) 334-344
11. Pearl, J.: Probabilistic Reasoning in Intelligent Systems: Networks of Plausible Inference. Morgan Kaufmann Publishers (1988)
12. Pelikan, M., Goldberg, D.: Research on the Bayesian Optimization Algorithms. IlliGAL Report No 200010, University of Illinois (2000)